\documentclass[sigconf]{acmart}
\newcommand{\ie}{\emph{i.e.~}}
\newcommand{\eg}{\emph{e.g.~}}
\newcommand{\etal}{\textit{et al.}}
\newcommand{\etc}{\emph{etc.}}

\AtBeginDocument{%
  \providecommand\BibTeX{{%
    \normalfont B\kern-0.5em{\scshape i\kern-0.25em b}\kern-0.8em\TeX}}}

\copyrightyear{2022}
\acmYear{2022}
\setcopyright{acmcopyright}
\acmConference[MM '22] {Proceedings of the 30th ACM International Conference on Multimedia}{October 10--14, 2022}{Lisboa, Portugal.}
\acmBooktitle{Proceedings of the 30th ACM International Conference on Multimedia (MM '22), Oct. 10--14, 2022, Lisboa, Portugal}
\acmPrice{15.00}
\acmISBN{978-1-4503-9203-7/22/10}
\acmDOI{10.1145/3503161.3548421}

\acmSubmissionID{3159}

\begin{document}

\title{High-Quality 3D Face Reconstruction with \\ Affine Convolutional Networks}

\author{Zhiqian Lin}
\authornote{Equal contribution}

\affiliation{%
\country{China,} \institution{Zhejiang University}
  }
\email{zhiqian_joy@zju.edu.cn}

\author{Jiangke Lin}
\authornotemark[1]
\affiliation{%
\country{China,} \institution{NetEase Fuxi AI Lab}
}
\email{linjiangke@corp.netease.com}

\author{Lincheng Li}
\affiliation{%
\country{China,} \institution{NetEase Fuxi AI Lab}}
\email{lilincheng@corp.netease.com}

\author{Yi Yuan}
\affiliation{%
\country{China,} \institution{NetEase Fuxi AI Lab}} 
\email{yuanyi@corp.netease.com}

\author{Zhengxia Zou}
\authornote{Corresponding author: zhengxiazou@buaa.edu.cn}
\affiliation{%
\country{China,} \institution{Beihang University}}
\email{zhengxiazou@buaa.edu.cn}

\renewcommand{\shortauthors}{Zhiqian Lin et al.}

\begin{abstract}
Recent works based on convolutional encoder-decoder architecture and 3DMM parameterization have shown great potential for canonical view reconstruction from a single input image.
Conventional CNN architectures benefit from exploiting the spatial correspondence between the input and output pixels. However, in 3D face reconstruction, the spatial misalignment between the input image (e.g. face) and the canonical/UV output makes the feature encoding-decoding process quite challenging. In this paper, to tackle this problem, we propose a new network architecture, namely the Affine Convolution Networks, which enables CNN based approaches to handle spatially non-corresponding input and output images and maintain high-fidelity quality output at the same time. In our method, an affine transformation matrix is learned from the affine convolution layer for each spatial location of the feature maps. In addition, we represent 3D human heads in UV space with multiple components, including diffuse maps for texture representation, position maps for geometry representation, and light maps for recovering more complex lighting conditions in the real world. All the components can be trained without any manual annotations. Our method is parametric-free and can generate high-quality UV maps at resolution of $512 \times 512$ pixels, while previous approaches normally generate $256 \times 256$ pixels or smaller. Our code will be released once the paper got accepted.

\end{abstract}



\begin{teaserfigure}
  \centering
  \includegraphics[width=0.98\linewidth]{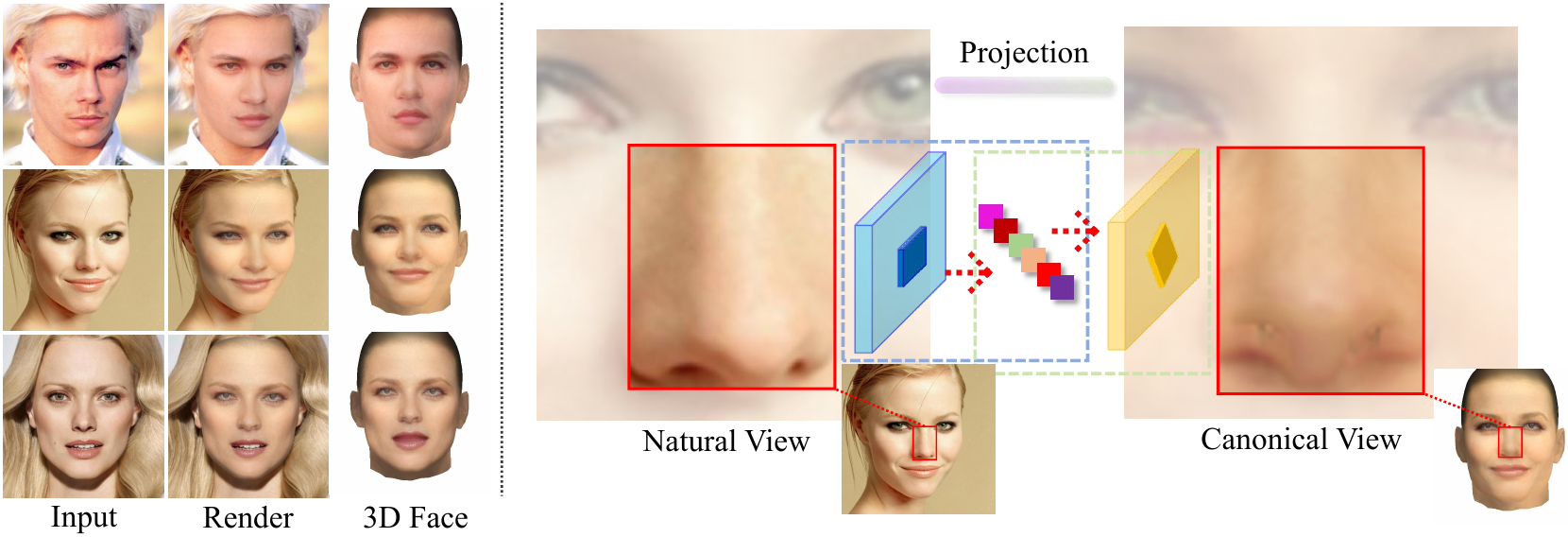}
  \caption{High-Quality 3D face reconstruction results from single in the wild images (left). The spatial misalignment between the natural view input and the canonical view output can be well handled with the proposed method (right).}
  \label{fig:contribution}
\end{teaserfigure}

\maketitle

\section{Introduction}

3D face/head reconstruction is an important problem in computer vision and graphics with a wide range of multimedia applications, such as avatar creation for metaverse or video games, virtual social interaction, plastic and cosmetology, \etc. Monocular 3D face reconstruction from in-the-wild images is challenging due to its ill-posed nature. 

In recent 3D face reconstruction literature~\cite{deng2019accurate,genova2018unsupervised,wu2019mvf,wu2020unsupervised,tewari2021learning,zhang2021learning}, encoder-decoder network architectures are commonly used where these methods first encode the input images into latent vectors, and then decode them into output geometry or textures, such as albedo, depth, and other information in canonical view~\cite{wu2020unsupervised,zhang2021learning}. On the one hand, recent studies on pixel-wise image prediction or image generation have shown the effectiveness of UNet-like~\cite{ronneberger2015u} architectures in generating high-resolution outputs. On the other hand, in 3D face reconstruction tasks, the input images are usually not spatially aligned with the outputs in canonical view or UV spaces, particularly those in-the-wild facial images. This causes a problem that the skip connections - a technique that is commonly used in UNet-like architectures for improving output fidelity, cannot be well applied to the face reconstruction due to the spatial misalignment issue. A dilemma is usually encountered in previous encoder-decoder based approaches when constructing high-resolution UV images: it is difficult to produce high-fidelity outputs while making full use of the spatial correspondence in convolutional networks. Therefore, current 3D face reconstruction methods usually encode the input images to a single or multiple latent vectors and avoid using any skip connections between the encoder and decoder layers. This design inevitably causes losses of reconstruction details.

To tackle the above problems, we proposed a new network architecture, namely the affine convolution networks for high-fidelity 3D face reconstruction. At the same time, by taking advantage of differentiable rendering, our method requires no manual annotation effort for training. We build the backbone of our networks based on the standard UNet~\cite{ronneberger2015u} architecture with minimal modifications. In our method, we design a new layer called the affine convolution layer, which makes the convolution capable of wrapping feature maps so that the input and output can be aligned when they are in different spatial transformations. With such a design, the skip connections can be well applied to our method to improve the resolution and fidelity of the reconstruction outputs. Our method takes a single input face image and predicts diffuse maps, position maps~\cite{feng2018joint} and light maps that recover full-stack information of the 3D face and lighting conditions, and finally achieves state-of-the-art reconstruction results. 
Our method is parametric-free and we propose to represent the 3D faces as well as the light condition fully in UV space. 
In this paper, we refer to our method as AffUNet for convenience.

Our contributions are summarized as follows:
\begin{itemize}
  \item We propose AffUNet - a new method for single image high-fidelity 3D face reconstruction. By leveraging differentiable rendering and 3DMM priors, the proposed method requires no manual annotation effort for training.
  \item A new neural network layer - affine convolution layer is proposed, which enables the network to generate high-resolution UV output while handling spatial misalignment at the same time. Our method operates at a resolution of $512 \times 512$ pixels, while previous methods only handle $256 \times 256$ or smaller images.
  \item Unlike previous methods that use a parameter vector to represent lights, we propose to represent the light conditions as well as the 3D faces fully in UV space that better handles complex illuminations in the real world.
\end{itemize}

\section{Related Works}
\subsection{3D Face Reconstruction}

Many recent works on 3D face reconstruction focus on using the 3D Morphable Model (3DMM)~\cite{blanz1999morphable,booth20163d,cao2013facewarehouse,gerig2018morphable,huber2016multiresolution,li2017learning,yang2020facescape} and deep neural networks. 3DMM aims to find a PCA model where the identity, expression, and texture of a 3D face are embedded in a low dimensional space with usually tens to hundreds of parameters. Recent 3DMM-based methods~\cite{deng2019accurate,richardson2017learning,sanyal2019learning,wu2019mvf,tewari2021learning} are either trained to directly predict 3DMM coefficients via neural networks or utilize non-linearity to improve the morphable model~\cite{tewari2017mofa,tewari2018self,tewari2021learning,genova2018unsupervised,tran2019towards}.

Thanks to the recent advances of differentiable rendering~\cite{genova2018unsupervised,kato2018renderer,liu2019soft}, the 3D faces reconstruction is now can be improved with self-supervised constraints~\cite{wu2019mvf,lin2020towards,lin2021meingame,zhou2019dense,zhu2020reda,lee2020uncertainty,luo2021normalized,deng2019accurate}. By leveraging the differentiable rendering, Wu~\etal~\cite{wu2020unsupervised} propose to reconstruct symmetric deformable 3D objects via fully unsupervised learning. They train several networks to predict a decomposition of the face into depth, albedo, illumination parameters, \etc. Zhang~\etal~\cite{zhang2021learning} extended and apply~\cite{wu2020unsupervised} to multiple in-the-wild images, generating better results.

\begin{figure*}
  \centering
   \includegraphics[width=0.95\linewidth]{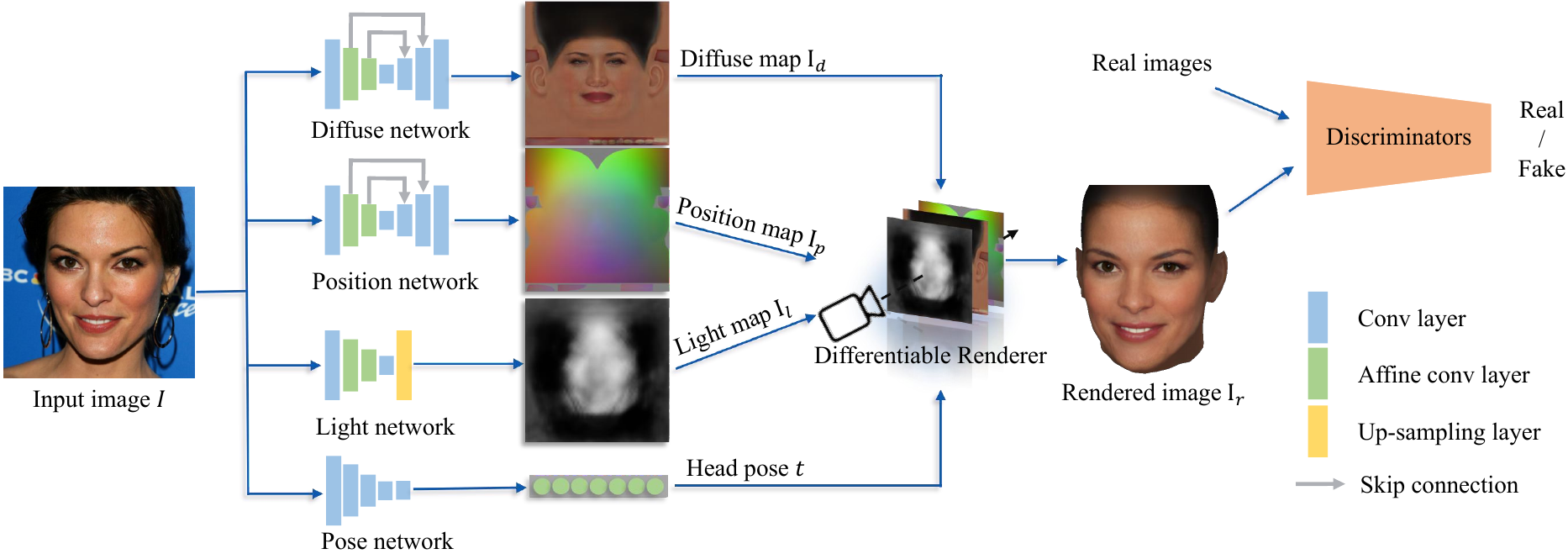}
   \caption{An overview of the proposed method. An input image $I$ is fed to four networks, \ie a diffuse network, a position network, a light network, and a pose network. The four networks produce the diffuse map $I_d$, position map $I_p$, light map $I_l$, and 3D head pose $t$, respectively. Then we render the reconstructed face image $I_r$ based on $I_d$, $I_p$, $I_l$, and $t$ with a differentiable renderer. Except for the position network that is trained with external 3DMM priors, the entire pipeline is trained in self-supervised manner by enforcing the reconstructed face image $I_r$ to be similar to the input $I$ as much as possible. The training does not require any manual annotations.}
   \label{fig:overview}
\end{figure*}

Despite the recent progress, texture fidelity is usually ignored in the previous 3DMM-based methods. Although the Generate Adversarial Networks (GAN)~\cite{goodfellow2014generative} are introduced to improve the visual quality of the generated texture maps recently~\cite{gecer2019ganfit,lattas2020avatarme,gecer2021ostec}, the inherent contradiction between spatial correspondence and misalignment of CNN network has not been addressed properly. Besides, the reconstruction results of most above methods can not effectively distinguish the foreground face and the background.

\subsection{Spatial Transformation in CNNs}

Spatial transformations and deformation modules in CNNs have brought increasing attention recently. The proposed AffUNet is highly related to these topics. 

Spatial Transformer Networks (STN)~\cite{jaderberg2015spatial} is one of the first to introduce the spatial transformation module in convolutional networks. STN transforms the feature maps according to the transformation parameters produced by an extra network. Although STN achieved success in small-scale images, STN is difficult to train since the transformation is globally applied in the entire feature maps. The inverse compositional STN~\cite{lin2017inverse} connects the core idea of the Lucas \& Kanade algorithm~\cite{lucas1981iterative} with STN and improves the transformation and the training efficiency.

Instead of learning global transformation, Dai~\etal~\cite{dai2017deformable} proposed to learn offsets for kernels in each pixel of feature maps. They refer to their method as the Deformable Convolutional Network (DCN). Zhu~\etal~\cite{zhu2019deformable} introduce additional deformable convolution layers with a modulation mechanism to improve DCN further. The main difference between the transformation in DCN and the proposed affine convolution layer is that the kernel offsets in DCN are learned in an element-wise fashion, while those in our method are regulated by affine transforms, which are naturally more suitable for representing image wrapping from 2D to UV spaces caused by the pose changes. We will explain the details in Sec.~\ref{sec:aff_conv}.

\section{Methodology}

Fig.~\ref{fig:overview} shows an overview of the proposed method. There are four basic network components in our method, a diffuse network, a position network, a light network, and a pose network. Given an input image $I$, the proposed networks jointly predict the position maps, diffuse maps, light maps, and pose parameters with the above four networks, respectively. All the outputs are represented in UV space except for the head poses. To close the training loop, we also introduce a differentiable renderer, which takes in those predicted components and generates a rendered image $I_r$. We train the networks to force the rendered result $I_r$ and the input $I$ as similar as possible. 

\subsection{Overview Architecture}

We build each component of our networks based on the well-known UNet architecture~\cite{ronneberger2015u}. As shown in Fig.~\ref{fig:overview}, we use two AffUNets to predict the diffuse map and the position map from the input image. To predict the light maps, we adopt an encoder-only network with affine convolution inside. The pose vector is predicted by a normal encoder-only network. Unlike other methods using a group of parameters to represent lighting (\eg directional light or point light or spherical harmonics light), we train the network to predict 2D light maps to represent complex lighting conditions. We experimentally find that 1) using colored (RGB channels) light maps, and 2) using light maps of the same (high) resolution as diffuse maps, will cause the facial textures (such as pores and moles) to be wrongly separated into light maps. Thus, we use single channel (\ie grayscale) images to store light maps. Depending on the application scenario, one can also train the network to predict three extra (RGB) values as the global illumination color, and then apply them to the light map. In addition to the diffuse maps, position maps, and light maps, we also predict the pose parameters $t$ of the head in 3D space, including the rotation and translation. The network structure is similar to the one used for predicting light maps, but we replace the last layer with a convolution layer with the kernel size equal to the feature map to produce a vector of $1\times 1$ spatial of the pose parameters.

In our method, the 3D faces are fully represented in UV spaces, including diffuse maps and position maps. 2D light maps are also used to present illumination. The position map is first introduced by Feng~\etal~\cite{feng2018joint}, which records the 3D shape of a complete face in UV space. Unlike the unsupervised method~\cite{wu2020unsupervised} that uses a depth map to represent a 3D shape that is hard to distinguish between foreground and background, the UV position maps we utilized can represent a complete face or head, which is more feasible for practical usage.

\subsection{Affine Convolution}\label{sec:aff_conv}

\begin{figure}
  \centering
   \includegraphics[width=\linewidth]{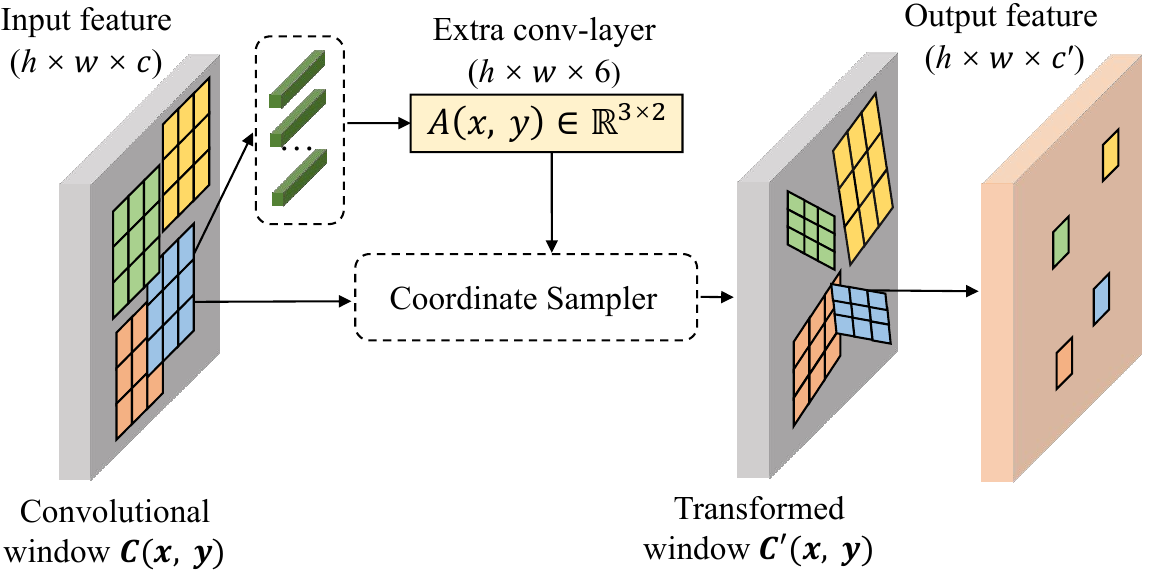}
   \caption{An illustration of the proposed affine convolution. The kernel size is set to $3 \times 3$ in this figure.}
   \label{fig:aff_conv}
\end{figure}

The main idea behind the proposed affine convolution layer is to locally learn a positional-wise affine wrapping transformation on the feature maps. Given a group of feature maps with three dimensions (\ie width, height, and channel), the affine transform is only applied to the spatial dimension and remains the same across the channel dimension. Compared to a standard convolution layer, the proposed affine convolution introduces two more computational steps: 1) predicting the affine transform matrix at each pixel location; 2) transforming the convolutional kernel according to the affine matrix, and sampling from the feature maps to complete the convolution operation. In Fig.~\ref{fig:aff_conv}, we illustrate the newly introduced affine convolution layer.

To introduce the proposed affine convolution layer, we first recap the standard convolutional layer in neural networks. For a $3 \times 3$ convolution kernel at coordinate $(x, y)$, the standard convolution will register a coordinate set of a $3 \times 3$ window
$$
  C(x,y) = \{(x-1, y-1),(x, y-1),...,(x+1, y+1)\},
$$
where $x$, $y$ are the coordinates of the pixel. Then the inner product between the $c(x,y)$ and the convolution kernel will be performed to produce the convolutional response at the location $(x, y)$.

In the proposed affine convolution layer, an affine transformation will be applied to the kernel before the inner production is performed. To predict the positional-wise affine transformation matrices $A$, an extra convolution layer is introduced. For the window location $C(x,y)$, we represent the transformation parameters $A(x,y)$ as follows:
\begin{equation}
    A(x,y) = \mathcal{F}(M_{C(x,y)})
\end{equation}
where $\mathcal{F}$ are the extra convolution layers and $M_{C(x,y)}$ are the features of window $C(x,y)$.

Given the predicted affine parameters $A(x,y)$ and the coordinate set $C(x,y)$, the new coordinates $C^\prime(x,y)$ for the kernel are represented as the matrix production of $[C(x,y) \quad 1]$ and $A(x,y)$:
\begin{equation}
  C^\prime(x,y) = [C(x,y) \ \ 1] \ A(x,y).
\end{equation}
For a $3\times 3$ kernel, the shape of $C^\prime(x,y)$, $ [C(x,y) \ \ 1]$ and $A(x,y)$ will thus become [9, 2], [9, 3] and [3, 2], respectively. Since the new coordinates $C^\prime(x,y)$ are not always integers, the interpolation operation will also be needed. We follow the sampling method in DCN~\cite{dai2017deformable} and the whole affine convolution is differentiable.

\subsection{Loss Functions}

\subsubsection{Reconstruction Loss}

The most straightforward way to design loss functions for the close-loop reconstruction task is to use pixel-wise L1 loss between the input and the rendered output. Given two images $I$ and $I_r$, the pixel-wise L1 loss is defined as follows:
\begin{equation}
  \mathcal{L}_{l1} (I, I_r) = \|I-I_r\|_1.
\end{equation}

However, we find it is difficult to directly predict the position map from an input image without any priors. We, therefore, create a set of ``pseudo ground truth'' of position maps to improve the training. The ``pseudo ground truth'' is created with an external 3DMM model, where we unwrap and interpolate the 3DMM model into UV space, and create the reference UV position maps. In this paper, we chose a pre-trained Deep3DFR~\cite{deng2019accurate} for generating the ``pseudo ground truth''.
Therefore, the L1 losses are used on both the rendered output and the predicted position maps. The final reconstruction loss is defined as:
\begin{equation}
  \mathcal{L}_{rec} = \lambda_{l1_d}\mathcal{L}_{l1}(I, I_r) + \lambda_{l1_p}\mathcal{L}_{l1}(I_p, I_p^\prime),
\end{equation}

where $I$ is the input image, $I_r$ is the rendered image, $I_p$ is the output position map, and $I_p^\prime$ is the position map created by the 3DMM mesh, and the $\lambda$s are the factors for balancing the loss terms. Except for the position network trained with external 3DMM priors, the entire pipeline is trained in a self-supervised manner.

\subsubsection{Perceptual Loss}

We apply another commonly used loss - perceptual loss, to regulate the distance between the rendered output and the input. The distance is computed in the feature space of a pre-trained VGG19 network~\cite{simonyan2014very}. For the two images $I$ and $I_r$, the perceptual loss is defined as follows:

\begin{equation}
  \mathcal{L}_{perc}(I, I_r) = \lambda_{perc} \sum_{i=1}^{N} \frac{1}{M_i} 
    \begin{Vmatrix}
    F^{(i)}(I) - F^{(i)}(I_r)
    \end{Vmatrix}_1,
\end{equation}
where $F^{(i)}$ denotes the $i$-th layer with $M_i$ elements of the VGG19 network. We use the 1st, 3rd, 5th, 9th, and 13th convolution layers for computing the loss.

\subsubsection{Symmetry Loss}

Since we represent 3D faces in the UV space, symmetry loss functions can be easily applied. For human faces, the diffuse maps are symmetric structures in principle. Thus, we penalize the differences between the Gaussian blurred diffuse map and its horizontal flip:
\begin{equation}
  \mathcal{L}_{sym}(I_r) = \lambda_{sym} \mathcal{L}_{l1}(\mathcal{G}(I_r), \mathcal{G}(\hat{I}_r)),
\end{equation}
where $\hat{I}_r$ is the horizontal flip of $I_r$, and $\mathcal{G}$ is the Gaussian blur function. We do not apply symmetry loss on the position maps and the light maps since they are not necessarily symmetrical.

\subsubsection{Regularization Losses}

We also deploy additional regularization losses on the diffuse maps and light maps, such as standard deviation loss and total variation loss.

The motivation behind applying the standard deviation loss is to make the skin tone to be similar across the entire face. We define this loss term as follows:
\begin{equation}
  \mathcal{L}_{std}(I_r) = \lambda_{std} \sqrt{\frac{1}{|\mathcal{M}_{skin}|}\sum_{i \in \mathcal{M}_{skin}}(\mathcal{G}(I)_i - \bar{I}_r)^2},
\end{equation}
where $\mathcal{M}_{skin}$ donates the mask of the skin region, $\bar{I}_r$ is the mean color of the skin regions.

For the light map, we assume the light should not be similar across the face, but should be close to neighboring pixels. Thus, the total variation loss is applied to the light map and is defined as follows:
\begin{equation}
  \mathcal{L}_{tv}(I_l) = \lambda_{tv} \sum_{x,y} |g_x(x,y)| + |g_y(x,y)|,
\end{equation}
where $g_x(x,y)$ and $g_y(x,y)$ are the image gradient of $I_l$ along the horizontal and vertical directions. $I_l$ is the predicted light map. $x, y$ are the pixel coordinates.

\subsubsection{Adversarial Losses.}

To generate high resolution ($512 \times 512$ pixels) and photo-realistic outputs, we adopt multi-scale discriminators~\cite{wang2018high} with adversarial training. We deploy 3 discriminators (\ie $D_1$, $D_2$, $D_3$) to differentiate real and the overlaid rendered face image at 3 different scales. The adversarial training loss is written as follows:

\begin{equation}
  \min_{G} \max_{D_1,D_2,D_3} \lambda_{adv} \sum_{k=1,2,3} \mathcal{L}_{GAN}(G, D_k),
\end{equation}
where $G$ is the collection of networks for generating diffuse maps, position maps, light maps, and 3D poses. $\mathcal{L}_{GAN}$ is the vanilla min-max loss of the GAN model.

\subsection{Implementation Details}
\subsubsection{Auxiliary Diffusion Losses}

Our method can be applied to any 3D face reconstruction application with a different texture style or different face/head topology, only a few target-style textures or a template face/head model are required. Here we choose a game environment as a test bed and evaluate our method for game character reconstruction. To make generated result compatible with character style of target game, we optionally adopt a few auxiliary human-annotated ground truths of diffuse maps for training.

During the training, all our networks components are updated in a self-supervised manner via differentiable rendering in most cases, but for those input images with stylized ground truth diffuse maps, an additional L1 loss is computed between the output diffuse map $I_d$ and ground truth $I_d^\prime$. A small number (about 2,000) of game-style diffuse maps can help the network to learn to generate reasonable style (\eg out-painting non-visible areas). The generated face model can be thus directly loaded into apps/games.

\subsubsection{Head Mesh and Position Map}
\label{sec:head_mesh_and_pos_map}

\begin{figure}[t]
  \centering
  \includegraphics[width=\linewidth]{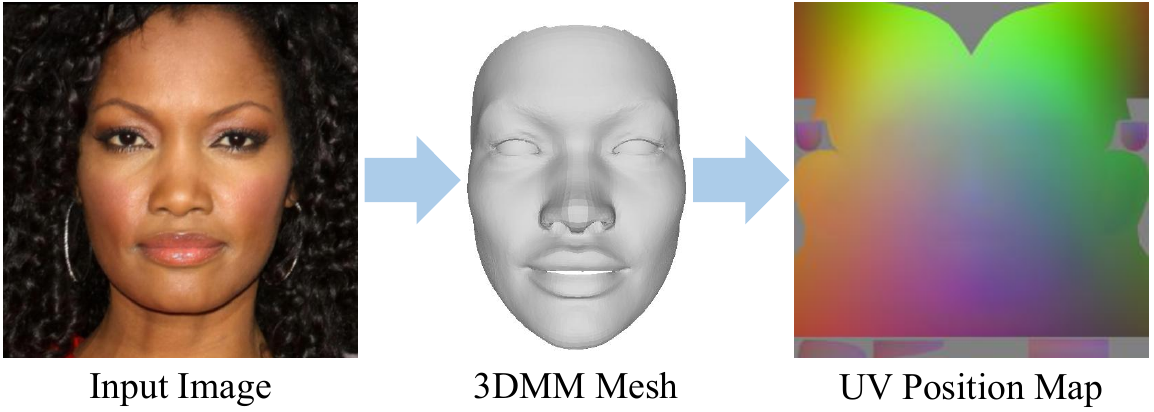}
  \caption{An illustration of the UV position map creation process.}
  \label{fig:position_map}
\end{figure}

The head mesh (with UV mapping correspondence) used as our template mesh~\cite{lin2021meingame} is from the PC game  \href{https://n.163.com}{``Justice''}. Note this template mesh is for in-game applications and is different from any other 3DMM model. To create pseudo ground truth for position maps, we 1) produce a 3DMM face mesh by~\cite{deng2019accurate} (\href{https://github.com/microsoft/Deep3DFaceReconstruction}{link}, under the \href{https://github.com/microsoft/Deep3DFaceReconstruction/blob/master/LICENSE}{MIT License}); 2) transform the shape to the template mesh~\cite{lin2021meingame}; 3) wrap and interpolate the head mesh to UV space according to the UV mapping correspondence. An illustration of this pipeline is shown in Fig.~\ref{fig:position_map}.

\subsubsection{Three-stage Training}

A three-stage training pipeline is proposed to stabilize training. The first two stages are used for training warm-up, and the last stage is for the joint training of all network components. We find that by training the geometry and texture parts separately during the warm-up phase, our networks can converge easier. The details of the three-stage training is given as follows:

\begin{itemize}
  \item Train the position network with $\mathcal{L}_{rec}$ first, by minimizing the differences between the output position map and its pseudo ground truth.
  \item Train the light network, diffuse network, and pose network with the following losses: $\mathcal{L}_{l1}(I,I_r)$, $\mathcal{L}_{perc}(I,I_r)$, $\mathcal{L}_{sym}(I_r)$, $\mathcal{L}_{std}(I_r)$, and $\mathcal{L}_{tv}(I_l)$. For the cases with paired ground truth diffuse maps, the losses $\mathcal{L}_{l1}(I_d,I_d^\prime)$ and $\mathcal{L}_{perc}(I_d,I_d^\prime)$ are also involved.
  \item After the above two stages of warm-up training, we jointly fine-tune all networks with all loss terms, also including the adversarial losses.
\end{itemize}

\subsubsection{Network Configurations}

In the proposed AffUNet, we replace some of the convolution layers of the UNet with affine convolution layers. Specifically, an AffUNet consists of 11 convolution layers. The first layer is a vanilla convolution layer with stride size = 1, and the second to fourth layers are affine convolution with stride size = 2. The followings are 7 vanilla convolution layers with stride size = 1. The decoder part consists of 3 upsampling layers. Unlike other methods that encode input images to latent vectors, which may cause loss of details, the AffUNet encodes inputs to feature maps and uses skip connections to produce high-resolution outputs, which preserve details of inputs. For the lighting branch, we down-sample the input image to a size of $1/8^2$ and then upsampling it by using bilinear interpolation. 

We conduct our experiment on the CelebA HQ~\cite{karras2018progressive} dataset (\href{https://github.com/tkarras/progressive_growing_of_gans}{link}, under the \href{https://github.com/tkarras/progressive_growing_of_gans/blob/master/LICENSE.txt}{CC BY-NC 4.0 License}). The resolution of 2D images and UV maps are both set to $512 \times 512$ pixels, while light maps are set to $64 \times 64$ pixels. When predicting the position maps, diffuse maps, light maps, and head pose, we down-sample the input image 3 times with three affine convolution layers, where the stride size is set to 2 for each of them. The template mesh and 2,000 auxiliary ground truth diffuse maps we use are from~\cite{lin2021meingame} (\href{https://github.com/FuxiCV/MeInGame}{link}, under the \href{https://github.com/FuxiCV/MeInGame/blob/master/LICENSE}{MIT License}).

We use the Adam optimizer for training and the learning rate is set to 0.001 with $\beta_1=0.9$ and $\beta_2=0.999$. We train our networks for 20 epochs in stage 1, 30 epochs in stage 2 and 40 epochs in the final fine-tune stage, respectively. We use grid-search on the loss weights from 0.001 to 10, and we select the best configuration based on the validation loss as well as the visual quality. The weights of loss terms are finally set as follows: $\lambda_{l1_d}=1$, $\lambda_{l1_p}=3$, $\lambda_{perc}=1$, $\lambda_{std}=1$, $\lambda_{tv}=0.3$, $\lambda_{sym}=0.3$, and $\lambda_{adv}=0.01$.

We run our experiments on an Intel i7 CPU and an NVIDIA 3090 GPU, and adopt PyTorch3D~\cite{johnson2020accelerating} as our differentiable renderer. During the inference time, given an input image of size $512 \times 512$, the whole networks take about 0.02s to produce the diffuse map and the position map with $512 \times 512$ pixels.

\begin{figure*}[t]
  \centering
   \includegraphics[width=0.8\linewidth]{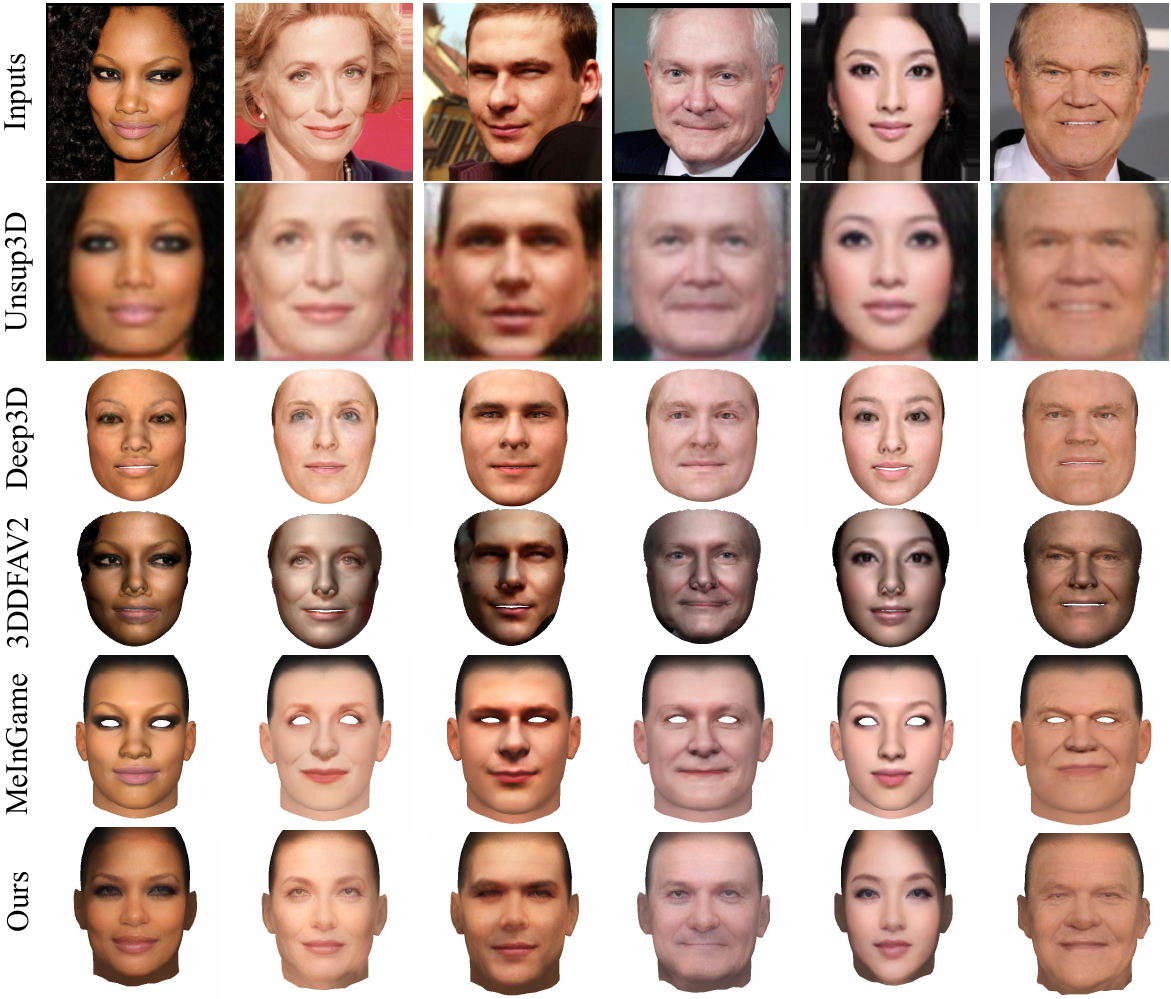}   \caption{Comparison of different methods: Unsup3D~\cite{wu2020unsupervised}, Deep3D~\cite{deng2019accurate}, 3DDFAV2~\cite{guo2020towards}, MeInGame~\cite{lin2021meingame}, and ours. All the results are rendered with their own predicted lights.}
   \label{fig:compare}
\end{figure*}
\section{Experimental Results}

\subsection{Qualitative Comparison}

Fig.~\ref{fig:compare} shows some comparison results between the 3DMM based methods and the proposed method. Unsup3D~\cite{wu2020unsupervised} is an unsupervised learning method, but it can only handle low-resolution images (\ie  $64 \times 64$ pixels), as reported in their paper, which also fails to recover details. 
However, ~\cite{wu2020unsupervised,zhang2021learning} produce depth maps to represent 3D face shapes, which are hard to differentiate occlusions from the background, and they are not in uniformed topology meshes, making them inflexible for practical usages. Besides, the resolution of $256 \times 256$ pixels ~\cite{zhang2021learning} is still lower than ours.
We can see that since Deep3DFR~\cite{deng2019accurate} represents the textures by a 3DMM texture PCA model, it cannot produce the details (\eg  wrinkles and pores) very well. 3DDFAV2~\cite{deng2019accurate} produces robust results to changes in occlusion, lighting and pose. However, the vertices are not aligned with the texture very well. MeInGame~\cite{lin2021meingame}, on the other hand, produces high-resolution texture maps and uniformed topology meshes. It unwraps the input images to UV space first, then translates them to diffuse maps, removing lights and occlusions. However, it tends to generate smooth results with a plastic-feeling effect (\eg the wrinkles in its second result).

All of the above comparison methods use a small number of parameters to represent the illumination. We can see whether it is directional lighting or spherical harmonic lighting, it is difficult to simulate the complex illumination of the human face very well. As a comparison, we propose to use light maps for illumination prediction. The results show that the network not only decouples the lighting from the facial textures, but also handles some occlusions (\eg hair) well very.

\begin{figure}
  \centering
  \includegraphics[width=0.938\linewidth]{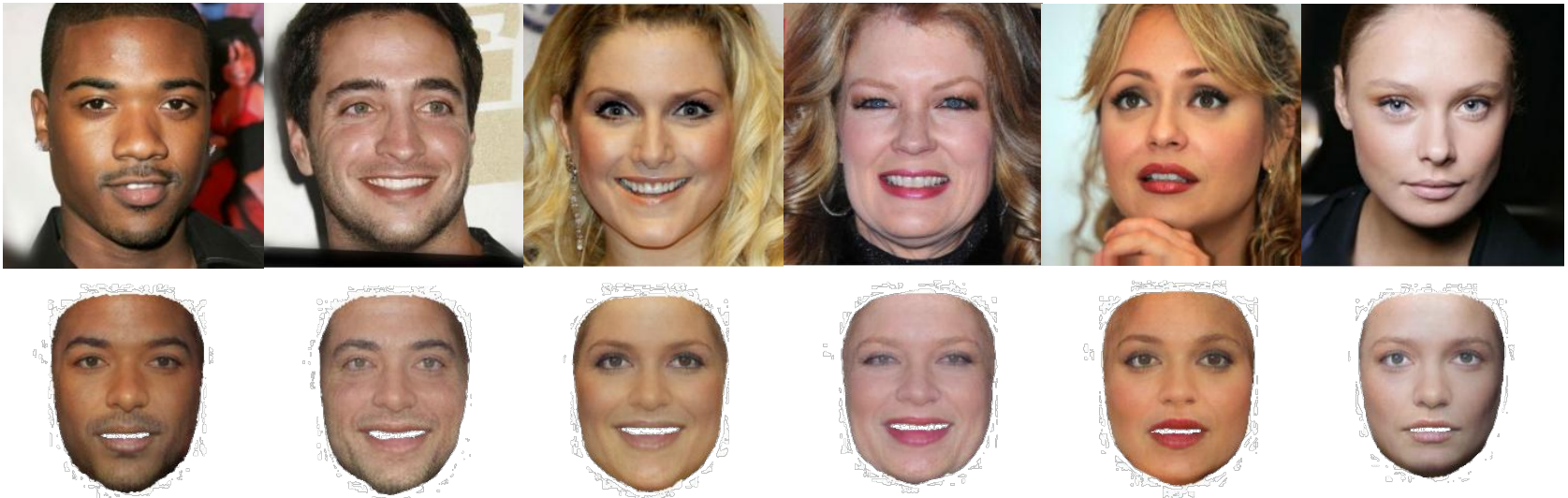}
  \caption{The reconstruction results of our proposed method using a different template mesh~\cite{paysan20093d}.}
  \label{fig:bfm_recomstruction}
\end{figure}

\begin{figure}
  \centering
   \includegraphics[width=\linewidth]{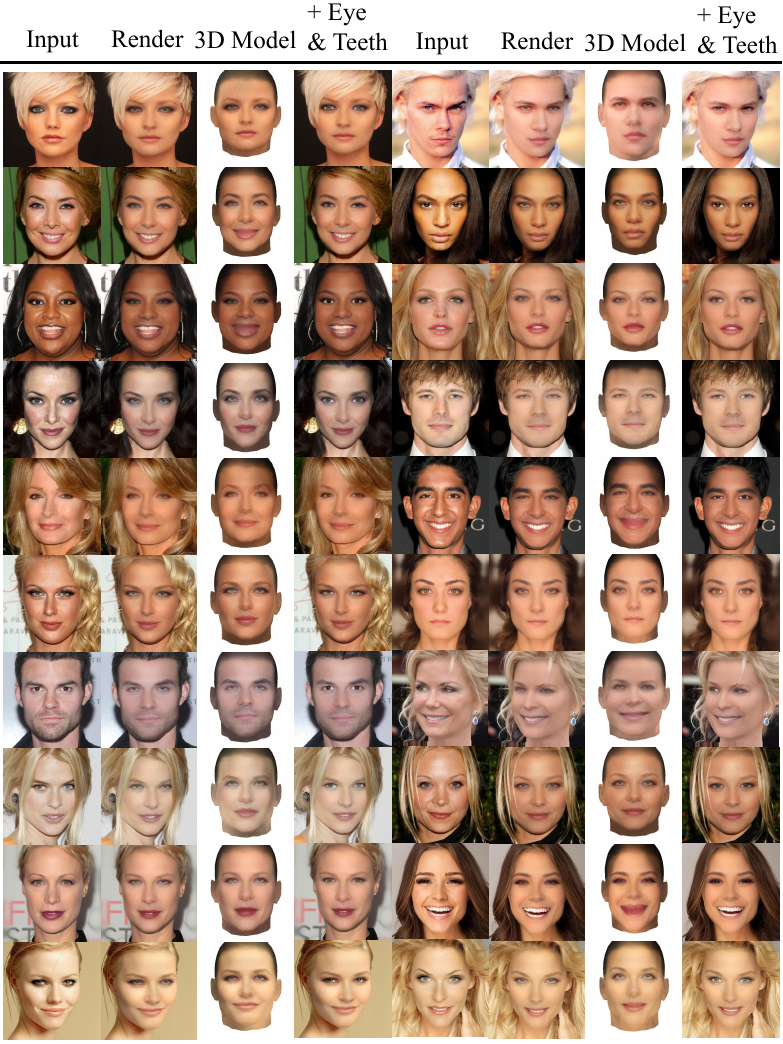}
   \caption{Quantitative results of reconstructed 3D models, rendering 3D model to the input image and rendered image with original eyes and teeth.}
   \label{fig:results}
\end{figure}

In addition, we also visualize some reconstructed results of the entire framework in Fig.~\ref{fig:results}, and the entire framework using BFM model~\cite{paysan20093d} as template mesh and training without target-style texture in Fig.~\ref{fig:bfm_recomstruction} to verify its universality and high-quality.

\begin{table}
\centering\setlength{\tabcolsep}{2.5pt}{
\begin{tabular}{c|cc}
\toprule
 Method & SIDE($\times 10^{-2}$) & MAD(deg.) \\
\hline
DECA\cite{DECA:Siggraph2021} & 2.486 & 53.61\\
3DDFAV2\cite{guo2020towards} & 1.731 & 46.79\\
Deep3DFR\cite{deng2019accurate} & 1.074 & 37.81\\
Unsup3D\cite{wu2020unsupervised} original & 1.423 & 23.55 \\
Unsup3D\cite{wu2020unsupervised} modified & 1.367 & 21.80 \\
Ours w/o affconv & 0.947 & 39.62 \\
Ours                    & \textbf{0.721} & \textbf{14.74} \\
\bottomrule
\end{tabular}}
\caption{Scale-Invariant Depth Errors (SIDE) and Mean Angle Deviation (MAD) of the reconstructed results evaluate on the BFM dataset~\cite{paysan20093d}. All methods are trained on real human face images in the wild.}
\label{tab:compare_geometry}
\end{table}

\begin{table}
\centering\setlength{\tabcolsep}{2.5pt}{
\begin{tabular}{c|ccc}
\toprule
 Method & Cosine Similarity & Top-1 & Top-5 \\
\hline
Deep3DFR\cite{deng2019accurate}
& 0.456 & 0.846 & 0.948 \\
Unsup3D\cite{wu2020unsupervised} original
& 0.393 & 0.667 & 0.897 \\
Unsup3D\cite{wu2020unsupervised} modified
& 0.429 & 0.744 & 0.949 \\
LAP\cite{zhang2021learning} & 0.471 & \textbf{0.923} & \textbf{1.0} \\
AvatarMe\cite{lattas2020avatarme} & 0.416 & 0.589 & 0.923 \\
Ours w/o affconv & 0.386 & 0.340 & 0.59 \\
Ours & \textbf{0.484} & 0.920 & \textbf{1.0} \\
\bottomrule
\end{tabular}}
\caption{Cosine similarity of feature embedding between input images and rendered images in canonical view. The features are extracted using a pre-trained face recognition network~\cite{wu2018light}. And the recall rate on the Top 1 / 5 neighbors.}
\label{tab:compare_texture}
\end{table}

\subsection{Quantitative Comparison}

We quantitatively compare our method with other methods in both geometry and texture quality. For geometry comparison, we follow the evaluation metrics in ~\cite{wu2020unsupervised,zhang2021learning}, and compute the scale-invariant depth error (SIDE) and mean angle deviation (MAD) between the predicted results and the ground truth of the BFM dataset~\cite{paysan20093d}. The methods in~\cite{deng2019accurate,guo2020towards,DECA:Siggraph2021,wu2020unsupervised} are used for comparison. The comparison results are shows in Tab.~\ref{tab:compare_geometry}.

To effectively evaluate the reconstructed 3D faces in facial feature similarity, it is more reasonable to evaluate the novel view rendered from the reconstructed 3D faces. Thus, we render the faces in canonical view (see Fig.~\ref{fig:compare} for examples) and compute the cosine distance of features extracted by a pre-trained face recognition network~\cite{wu2018light} between input images and rendered images. In addition, we also look up the nearest neighbor face images according to the cosine similarity. Top-1 and Top-5 show the recall rate on the nearest neighbor or the nearest 5 neighbors. All the scores can be found in Tab.~\ref{tab:compare_texture}.

\begin{figure}
  \centering
   \includegraphics[width=0.9\linewidth]{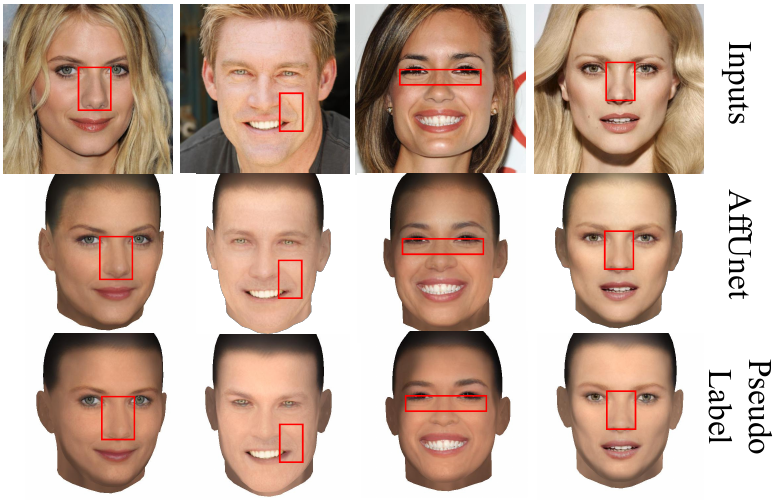}
   \caption{Comparison of our full implementation with predict position maps (the second row) and with pseudo ground truth generated by 3DMM-based method (the last row).}
   \label{fig:compare_pseudo}
\end{figure}

We also compare the results rendered based on our predicted position maps and those based on the ``pseudo ground truth'' created by the external 3DMM model~\cite{deng2019accurate}. In Fig.~\ref{fig:compare_pseudo}, we can see from the comparison results that although the position map branch is trained based on the 3DMM pseudo ground truth, our final reconstruction results are better than their ground truth in visual fidelity details.

\subsection{Ablation Study}

\begin{figure}
  \centering
   \includegraphics[width=0.9\linewidth]{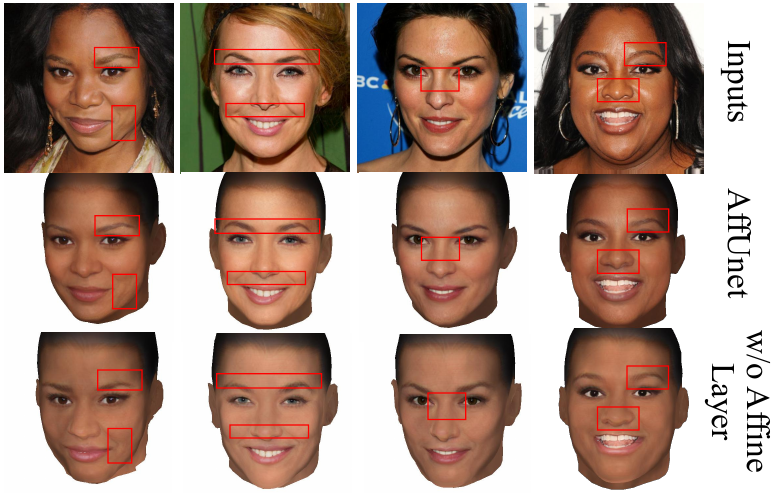}
   \caption{Ablation study on the affine convolution layer. Rows from top to bottom: input images, our full implementation, and our method w/o affine convolution layer.}
   \label{fig:3dmm_confirm_affconv}
\end{figure}

\noindent\textbf{Effectiveness of Affine Convolution.}
To verify the effectiveness of the proposed affine convolution layer, for qualitative comparison, we replace the affine convolution layer with a standard convolution layer and compare their reconstruction results. For a fair comparison, all the other configurations remain unchanged. The comparison results are visualized in Fig.~\ref{fig:3dmm_confirm_affconv}. The results without the affine convolution layer look blurry and lack reconstruction details compared to the complete method. We also quantitatively compare them in both geometry and texture quality in Tab.~\ref{tab:compare_geometry} and Tab.~\ref{tab:compare_texture}.

\begin{figure}
  \centering
   \includegraphics[width=\linewidth]{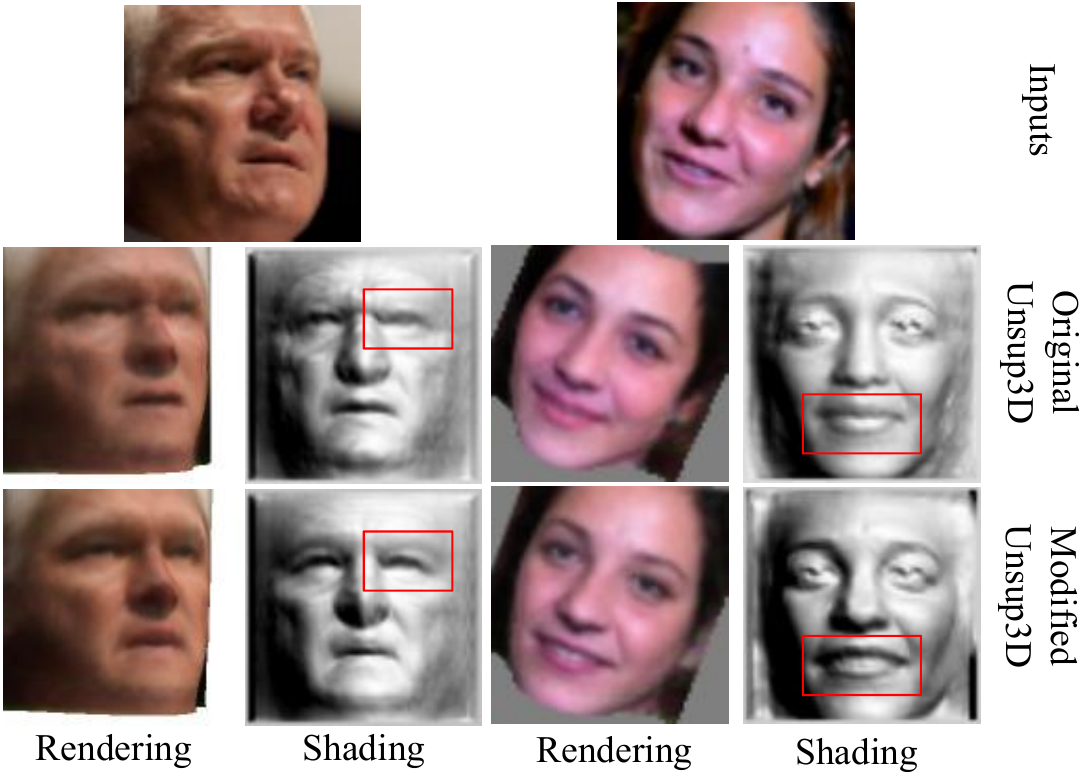}
   \caption{Ablation study of proposed affine convolution replaces on Unsup3D dataset~\cite{wu2020unsupervised}. The network with affine convolution performs better than plain version, particularly in details of lips and eyes. Best view in color and zoom in.}
   \label{fig:unsup3d_confirm_affconv}
\end{figure}

Not limited to our method, we also find that the proposed affine convolution can be applied to other face reconstruction methods, e.g., the unsupervised method~\cite{wu2020unsupervised}, and can help improve the reconstruction details. For a fair comparison, we replace two vanilla convolution layers in the network~\cite{wu2020unsupervised} with the proposed affine convolution layer, and train the original network and the modified one under the same configuration. The comparison results are shown in Fig.~\ref{fig:unsup3d_confirm_affconv}.

\begin{table}
\centering\setlength{\tabcolsep}{2.5pt}{
\begin{tabular}{c|ccc}
\toprule
    Method   & Cosine Similarity & Top-1 & Top-5 \\
\hline
w/o $\mathcal{L}_{perc}$ & 0.317 & 0.359 & 0.716 \\
w/o $\mathcal{L}_{sym}$  & 0.361 & 0.589 & 0.769 \\
w/o $\mathcal{L}_{std}$  & 0.362 & 0.641 & 0.846 \\
w/o $\mathcal{L}_{adv}$  & 0.447 & 0.794 & 0.872 \\
Ours full model & \textbf{0.484} & \textbf{0.920} & \textbf{1.0} \\
\bottomrule
\end{tabular}}
\caption{Ablation study of different loss terms: perception loss $\mathcal{L}_{perc}$, symmetric loss $\mathcal{L}_{sym}$, standard deviation loss $\mathcal{L}_{std}$, and the adversarial loss $\mathcal{L}_{adv}$.}
\label{tab:ablation_loss}
\end{table}

\begin{figure}
  \centering
  \includegraphics[width=\linewidth]{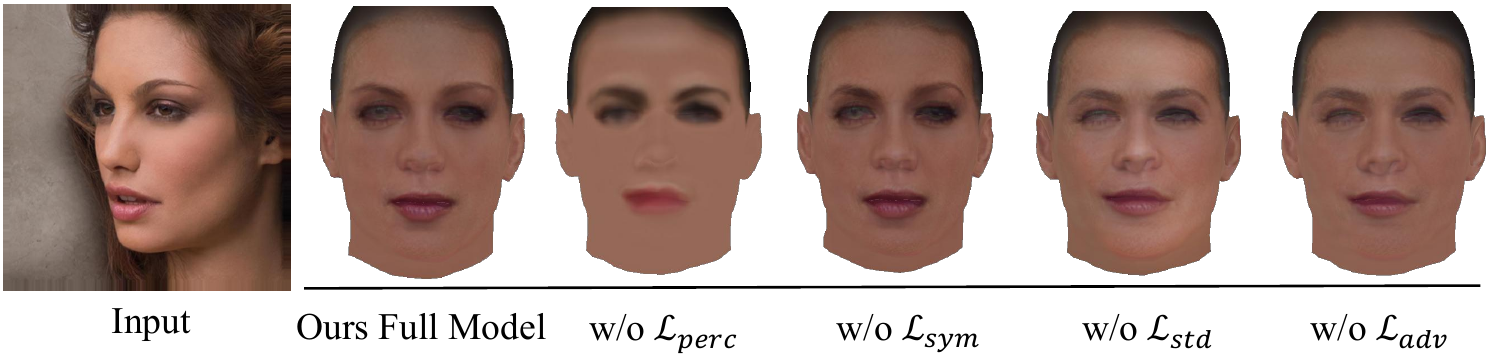}
  \caption{Ablation study of different loss terms. For a better comparison, the 3D heads are rendered in canonical view with only diffuse maps and \textbf{without} lighting.}
  \label{fig:loss_ablation}
\end{figure}

\noindent\textbf{Ablation on Different Losses.}
We then analyze the contributions of different loss terms we adopted. As shown in Tab.~\ref{tab:ablation_loss}, the network performance is influenced by each of the loss terms to some degree. The full implementation of our method achieves the highest score in terms of both cosine similarity and nearest neighbor recall accuracy (top-1 and top-5). From Fig.~\ref{fig:loss_ablation}, we can see that without the perceptual loss, the network produces texture maps with limited details - with only an average of the skin colors. The symmetric loss helps the network predicts symmetric faces, such as eyebrow shapes. The skin regularization loss penalizes the differences in skin tone across the whole face. Without it, the generated diffuse maps could be baked with lights. Meanwhile, the adversarial loss can further improve the fidelity of skin and facial features.

\subsection{Limitations}

A limitation of our method is that the training of our position network still relies on the inductive bias created by an external 3DMM-based method. Without the assistance of 3DMM, the shape prediction network may have difficulty in distinguishing between the face and background during the transformation from 2D image to UV space. 

Like other 3D face reconstruction methods, the performance of our approach is also limited by the image quality and distribution of the training set. Since most images of the CelebA dataset are Caucasian faces, the network may not be as accurate in predicting the diffuse color of Asian or African faces.

\section{Conclusion}

In this paper, we introduce a new approach called Affine Convolutional Networks for 3D face reconstruction from a single input face image. With the help of the proposed affine convolution layers, the network can operate at resolution $512 \times 512$ pixels, generating high-quality and photo-realistic 3D faces with rich facial fidelity. Extensive experiments suggest the effectiveness of the affine convolution layer and other components of our design. In our future work, we will include other transformations, \eg perspective transformation, to the convolution layers. More applications on other tasks such as view synthesis will also be explored. 

\bibliographystyle{ACM-Reference-Format}
\bibliography{sample-base}

\end{document}